\documentclass[10pt,twocolumn,letterpaper]{article}

\usepackage{cvpr}
\usepackage{times}
\usepackage{epsfig}
\usepackage{graphicx}
\usepackage{amsmath}
\usepackage{amssymb}
\usepackage{array}
\usepackage{caption}

\usepackage{multirow}
\usepackage{fancyhdr}
\usepackage{hyperref}
\usepackage{booktabs}

\usepackage{float}

\newcommand{\PreserveBackslash}[1]{\let\temp=\\#1\let\\=\temp}
\newcolumntype{C}[1]{>{\PreserveBackslash\centering}p{#1}}
\newcolumntype{R}[1]{>{\PreserveBackslash\raggedleft}p{#1}}
\newcolumntype{L}[1]{>{\PreserveBackslash\raggedright}p{#1}}

\cvprfinalcopy 

\def\cvprPaperID{****} 

\begin{document}

\title{BANet: Motion Forecasting with Boundary Aware Network}

\author{Chen Zhang\textsuperscript{1}
\quad
Honglin Sun\textsuperscript{1,2}
\quad 
Chen Chen\textsuperscript{1}
\quad
Yandong Guo\textsuperscript{1} 
\\
\textsuperscript{1}OPPO Research Institute.
\quad
\textsuperscript{2}Waseda University.
\\
{\tt\small \{zhangchen4, chenchen, guoyandong\}@oppo.com}
\\
{\tt\small hsun@akane.waseda.jp}
}

\twocolumn[{%
\renewcommand\twocolumn[1][]{#1}%
\maketitle
\begin{center}
    \centering
    \includegraphics[scale=0.54]{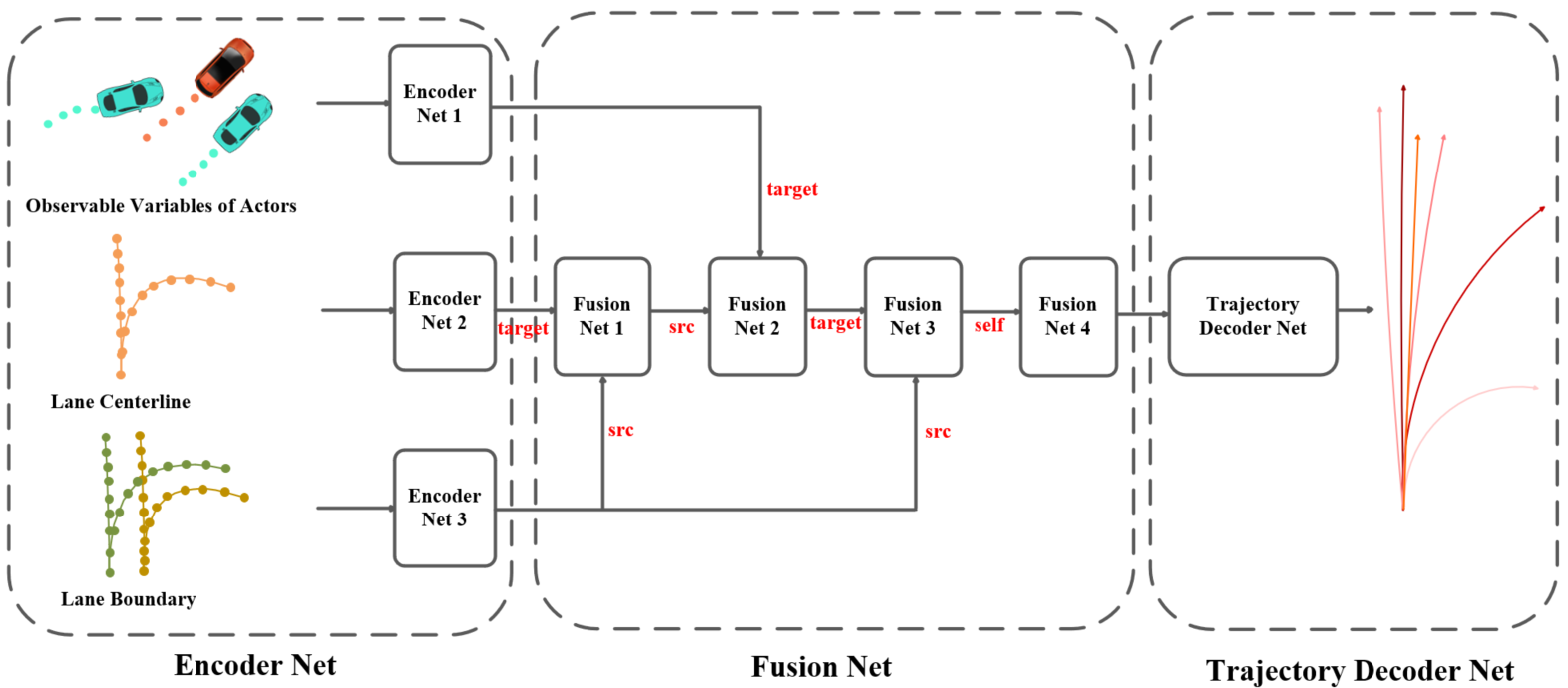}
    \captionof{figure}{The overall network structure of BANet.We use Encoder Net to extract the embeddings of the vectorized data. Then Fusion Net fuses the embedding features. Trajectory Decoder Net outputs the final predicted future trajectories.} \label{fig:overview}
\end{center}%

}]



\section{Method}
For motion forecasting task, methods using vectorized input have been shown to be more efficient than methods using rasterized input, and can get a better performance. Our motion forecasting model also uses vectorized data as input.

We call our motion forecasting model \textbf{BANet}, which means boundary-aware network, and it is a variant of LaneGCN\cite{liang2020learning}. We believe that only using lane centerline as input to get the embedding feature of vector map nodes is not enough. The lane centerline can only provide the topology of the lanes, and other elements of the vector map also contain rich information. For example, the lane boundary can provide traffic rule constraint information such as whether it is possible to change lanes which is very important. Therefore, we believe a motion forecasting model can get a better performance by encoding more vector map elements and do feature fusion on them. We report our results on the 2022 Argoverse2 Motion Forecasting challenge and rank 1st on the test leaderboard.

\subsection{Architecture}
Our motion forecasting model consists of three parts: Encoder Net, Fusion Net, and Trajectory Decoder Net. We use the Encoder Net to extract embedding from different vectorized data. And then we use the Fusion Net to fuse and exchange information between actors and vector map. Finally we use the Trajectory Decoder Net to decode multiple trajectories. The whole network is shown in Fig.\ref{fig:overview}.

\noindent{\textbf{Encoder Net:}} We divide the vectorized data into three parts: historical observable variables of actors, lane centerlines, and lane boundaries. Correspondingly, the Encoder Net also consists of three parts. For the historical observable variables of actors, including coordinate, heading angle, and velocity, we use 1D CNN combined with a max pooling layer to extract features. The coordinate, heading angle and velocity first pass through a 1D CNN resnet block respectively, and then the output features are added together and then pass through more 1D CNN resnet blocks and a FPN. For the lane centerlines and lane boundaries of the vector map, we use MLP to extract features respectively. The lane graph convolution structure proposed in LaneGCN\cite{liang2020learning} is an effective method to update information on the lane centerlines according to the topological relationships, but for different lane centerline nodes, the influence of each category of adjacency relationship is different. Therefore, we add learnable weights to lane graph convolution for each category of the adjacency relationship, thereby we can generate a set of different weights for different lane centerline nodes to represent the influence of the adjacency relationship of different categories. Experiments show that this improvement has positive benefits.

\noindent{\textbf{Fusion Net:}} As shown in the Fig.1, our Fusion Net consists of four sub-fusion blocks. First, the features of the lane boundaries are fused into the lane centerlines through the matching relationship between the lane boundary and the lane centerline, so that each node on the lane centerline can get the traffic rule constraints provided by the nearby boundary. Secondly, the features of the lane centerline are fused into the features of the actors. And then the features of the boundary are fused into the actors, so that the actors can know the nearby lane constraints. Finally, feature interactions are performed between actors in the scenario. Except for the first sub-fusion block, other sub-fusion blocks are implemented by the distance attention module which is proposed in LaneGCN\cite{liang2020learning}.

\noindent{\textbf{Trajectory Decoder Net}} As mentioned in many related works, the target point of the predicted trajectory contain most of the trajectory's intent information, so we divide the trajectory decoder net into two stages. In the first stage, we regress the target point. And in the second stage, we encode the target points regressed in the first stage and concatenate them with the features of the actors to decode the complete trajectory points. For each agent, we get six future trajectories with confidence.

\subsection{Loss Function}
Since we decompose the prediction task into two sub-tasks: target prediction and trajectory completion, the final loss consists of three parts: trajectory confidence loss, target regression loss, and trajectory regression loss. The calculation details are as follows.

\subsubsection{Trajectory confidence loss}
For the confidence loss, we dropped predictions with $\rm minFDE > 2\rm m$ during training, in which case, we consider the model does not yield a good prediction due to the lack of encoded information. 

We use a max entropy model to assign the ground truth confidence score to each trajectory to get the ground truth distribution.

\begin{equation}
    \mathbf{c(s)} = \frac{{\rm exp}(-\mathcal{D}(\mathbf{s}, \hat{\mathbf{s}}))}{\sum_{i=1}^{N}{{\rm exp}(-\mathcal{D}(\mathbf{s}_i, \hat{\mathbf{s}}))}},
    \label{cls}
\end{equation}

where $\mathbf{s}$ is predicted trajectories, $\hat{\mathbf{s}}$ is the ground truth trajectory. For each time step in the trajectory $\mathbf{s}_i$, the displacement error $\mathcal{D}(\mathbf{s}_i, \hat{\mathbf{s}}) = {\rm max}(\Vert s_{i,0} - \hat{s}_{i,0}\Vert _2, \Vert s_{i,1} - \hat{s}_{i,1}\Vert _2,..., \Vert s_{i,T} - \hat{s}_{i,T}\Vert _2)$

We utilize Kullback-Leibler Divergence to calculate the loss, aiming to make the predicted probability distribution close to the ground truth distribution.

\begin{equation}
    \mathcal{L}_{{\rm conf}} = \frac{1}{N}\sum_i^N\mathcal{L}_{{\rm KL}}(\mathbf{c}_i, \hat{\mathbf{c}}_i),
    \label{cls}
\end{equation}

where $\mathbf{c}$ and $\hat{\mathbf{c}}$ are the predicted confidence scores and the ground truth scores respectively. 

\subsubsection{Target regression loss}
For the target regression, we only regress target points of the actors that have observations at the last time step, ensuring that the task of the target prediction net is always to predict the coordinates on the same time step.

\begin{equation}
    \mathcal{L}_{{\rm target}} = \frac{1}{N}\sum_i^N\mathcal{L}_{{\rm reg}}(\mathbf{g}_i, \hat{\mathbf{g}}_i),
    \label{reg_target}
\end{equation}

where, $\mathcal{L}_{{\rm reg}}$ is the smooth $\ell_1$ loss over the offset between the predicted target points $\mathbf{g}$ and the ground truth target points $\hat{\mathbf{g}}$.

\subsubsection{Trajectory regression loss}
For the trajectory regression, we consider only the coordinates before the last time steps. The coordinate of each time step is denoted with $\mathbf{s}_t$. Similar to the target regression loss, we adopt the smooth $\ell_1$ loss.

\begin{equation}
    \mathcal{L}_{{\rm traj}} = \frac{1}{N(T-1)}\sum_i^N\sum_t^{T-1}\mathcal{L}_{{\rm reg}}(\mathbf{s}_{i,t}, \hat{\mathbf{s}}_{i,t})
    \label{reg}
\end{equation}

\subsubsection{Total loss}
Our total loss is a linear combination of the above loss terms. We divide the training process into two stages. The first stage is to train the network except the trajectory completion module. Since at the early stage of training, the trajectory completion module cannot obtain accurate target point features from the target prediction net. Thus, similarly, only the coordinates of the target points are considered for calculating the confidence loss at this stage.

\begin{table*}[t]
\centering
\caption{Experiments on the Argoverse1 dataset. The results are on the Argoverse1 testset.}
\label{tab:table1}
\scalebox{0.72}{
\begin{tabular}{cccccccccc}
\hline
{Method}                  &{brier-minFDE (K=6)} &{minFDE (K=6)} &{minFDE (K=1)} &{brier-minADE (K=6)} &{minADE (K=6)} &{minADE (K=1)} &{MR (K=6)} &{MR (K=1)}\\
\hline
LaneGCN                  & 2.056                     & 1.362               & 3.803                & -                 & 0.864                   & 1.717  &0.161 & 0.592\\
BANet(lite)            & 1.894                   & 1.242                & 3.538                & -                & 0.816                  & 1.621  &0.135 & 0.558\\

\hline
\end{tabular}}
\end{table*}

\begin{table*}[t]
\centering
\caption{Experiments on the Argoverse2 dataset. The results are on the Argoverse2 testset.}
\label{tab:table2}
\scalebox{0.7}{
\begin{tabular}{cccccccccc}
\hline
{Method}                  &{brier-minFDE (K=6)} &{minFDE (K=6)} &{minFDE (K=1)} &{brier-minADE (K=6)} &{minADE (K=6)} &{minADE (K=1)} &{MR (K=6)} &{MR (K=1)}\\
\hline
BANet(lite)                  & 2.1286                     & 1.4861              & 5.0032              & 2.4026               & 0.7662                   & 1.9457  &0.203& 0.6336\\
BANet            & 2.033                   & 1.3888               & 4.6997                & 2.3031              & 0.733                 &  1.84  &0.1797 & 0.6152\\
BANet+ensemble            & 1.9203                  & 1.3648              & 4.6065              & 2.1789               & 0.7075                 & 1.7927  &0.1864 & 0.6007\\

\hline
\end{tabular}}
\end{table*}

\begin{equation}
    \mathcal{L}_{{\rm S1}} = \mathcal{L}_{{\rm conf}} + \mathcal{L}_{{\rm target}}.
    \label{s1}
\end{equation}

In the second stage, we add the trajectory completion module to train the whole network.
\begin{equation}
    \mathcal{L}_{{\rm S2}} = \mathcal{L}_{{\rm conf}} + \mathcal{L}_{{\rm target}} + \mathcal{L}_{{\rm traj}}.
    \label{s2}
\end{equation}

\subsection{Implementation details}
Our model is trained on the training set with a batch size of 32. We adopt NAdam \cite{dozat2016incorporating} optimizer and cosine annealing with warm restart \cite{loshchilov2016sgdr} learning rate, making the learning rate periodically decay and restart between $1\times10^{-3}$ and $1\times10^{-5}$. Specifically, we let the learning rate restart at the epoch 6 for the first time, and double the period after each restart. The training procedure will last for 4 restart periods (90 epochs) in total, where the second training stage starts at the first restart. At the end of the last period, we continue training the model with the minimum learning rate ($1\times10^{-5}$) until epoch 100 to make the network stable.

\subsection{Ensemble}
Ensemble is an important trick to refine the final prediction results. We selected 7 sub-models with different learning rates and distance attention thresholds,  which means that for each agent,  we have 42 predicted future trajectories. Then we use k-means to cluster the 42 target points and set the number of cluster centers to six. The final six trajectories are the average values calculated by multiplying the original trajectories contained in the six clusters by the weight $\mathcal{W}$ respectively. It should be noted that the result of multiplying the confidence $\mathcal{\rm c}$ of each original trajectory by the weight $\mathcal{W}$, as the sampling weight of k-means, will get a better result than the general k-means clustering. The weight $\mathcal{W}$ is calculated according to the value of the brier-minFDE $\mathcal{\alpha}$ of the original models used for clustering, and the calculation formula is as follows:
\begin{equation}
    \mathcal{W}_{{\rm i}} = \mathcal{\rm c}_{{i}} * \frac{{\rm exp}(-\mathcal{\alpha}_{j})}{\sum_{j=1}^{N}{{\rm exp}(-\mathcal{\alpha}_{j}))}}
    \label{s2}
\end{equation}
$\mathcal{N}$ is the number of submodels used for clustering, and for each trajectory used for clustering, its corresponding alpha is the value of the brier-minFDE of the model to which it belongs.

\section{Experiments}
Because the performance validation of our BANet was initially performed on the Argoverse1 dataset\cite{Argoverse}, our experimental records consist of two stages,  the Argoverse1 dataset\cite{Argoverse} and the Argoverse2 dataset\cite{Argoverse2}. The experimental results on the testset of the Argoverse1 dataset\cite{Argoverse} are shown in Table \ref{tab:table1}, and the experimental results on the testset of the Argoverse2 dataset\cite{Argoverse2} are shown in Table \ref{tab:table2}.

Here we compare the performance of BANet and LaneGCN\cite{liang2020learning} on the testset of Argoverse1 dataset\cite{Argoverse}. Because the vector map of the Argoverse1 dataset\cite{Argoverse} does not contain lane boundary, and the number of observable variables for actors is also less than that of the Argoverse2 dataset\cite{Argoverse2}. Therefore, the BANet we tested on the Argoverse1 dataset\cite{Argoverse} does not contain boundary encoder and fusion, nor does its input contain heading angle and velocity. We call this model BANet(lite).

{
\bibliographystyle{ieee}
\bibliography{egpaper_final}

\begin{thebibliography}{1}\itemsep=-1pt

\bibitem{Argoverse}
M.-F. Chang, J.~W. Lambert, P.~Sangkloy, J.~Singh, S.~Bak, A.~Hartnett,
  D.~Wang, P.~Carr, S.~Lucey, D.~Ramanan, and J.~Hays.
\newblock Argoverse: 3d tracking and forecasting with rich maps.
\newblock In {\em Conference on Computer Vision and Pattern Recognition
  (CVPR)}, 2019.

\bibitem{dozat2016incorporating}
T.~Dozat.
\newblock Incorporating nesterov momentum into adam.
\newblock 2016.

\bibitem{liang2020learning}
M.~Liang, B.~Yang, R.~Hu, Y.~Chen, R.~Liao, S.~Feng, and R.~Urtasun.
\newblock Learning lane graph representations for motion forecasting.
\newblock In {\em European Conference on Computer Vision}, pages 541--556.
  Springer, 2020.

\bibitem{loshchilov2016sgdr}
I.~Loshchilov and F.~Hutter.
\newblock Sgdr: Stochastic gradient descent with warm restarts.
\newblock {\em arXiv preprint arXiv:1608.03983}, 2016.

\bibitem{Argoverse2}
B.~Wilson, W.~Qi, T.~Agarwal, J.~Lambert, J.~Singh, S.~Khandelwal, B.~Pan,
  R.~Kumar, A.~Hartnett, J.~K. Pontes, D.~Ramanan, P.~Carr, and J.~Hays.
\newblock Argoverse 2: Next generation datasets for self-driving perception and
  forecasting.
\newblock In {\em Proceedings of the Neural Information Processing Systems
  Track on Datasets and Benchmarks (NeurIPS Datasets and Benchmarks 2021)},
  2021.

\end{thebibliography}
}
\end{document}